\documentclass[10pt,twocolumn]{article}

\usepackage{times}
\usepackage{epsfig}
\usepackage{graphicx}
\usepackage{amsmath}
\usepackage{amssymb}
\usepackage{dsfont}
\usepackage{tabularx}
\usepackage{stfloats}
\graphicspath{{./images/}}

\usepackage[pagebackref=true,breaklinks=true,letterpaper=true,colorlinks,bookmarks=false]{hyperref}

\begin{document}

\title{Predicting Clinical Outcomes with Waveform LSTMs}

\author{Michael Charles Albada\\
        Microsoft\\
        {\tt\small malbada@microsoft.com}}

\maketitle

\begin{abstract}
   Data mining and machine learning hold great potential to enable health systems to systematically use data and analytics to identify inefficiencies and best practices that improve care and reduce costs. Waveform data offers particularly detailed information on how patient health evolves over time and has the potential to significantly improve prediction accuracy on multiple benchmarks, but has been widely under-utilized, largely because of the challenges in working with these large and complex datasets. This study evaluates the potential of leveraging clinical waveform data to improve prediction accuracy on a single benchmark task: the risk of mortality in the intensive care unit. We identify significant potential from this data, beating the existing baselines for both logistic regression and deep learning models. 
\end{abstract}

\section{Introduction}

Since 2010, US hospitals have been eligible for financial incentives through Medicare and Medicaid for adopting and using electronic health records (EHRs) in accordance with federal meaningful-use criteria. In 2015 [10], nearly all reported hospitals (96
 
The burden of care for critically ill patients is massive. For instance, in the United States, it currently accounts for nearly 1

Most of these ICU severity scores leverage models that impose stringent constraints on the relationship between explanatory variables and risk of death. Given that the true relationship between risk of mortality in the ICU and explanatory variables is unknown, we expect that prediction can be improved by using a machine learning algorithm to estimate risk of death without requiring any specification about the shape of the underlying relationship. Additionally, these models utilize patient demographics and physiological variables such as age, temperature, and heart rate collected within the initial 12 to 24 hours after ICU admission with the goal of measuring the ICU efficacy. 
 
To enhance the efficiencies of the score-based models, several customized models have been proposed. A. Dervishi, et al. [15], for example, introduces a model that leverages the cardiorespiratory arrest data to predict patient mortality. Other studies ([16][17][18][19][20]) preferred sophisticated data mining techniques such as random forest, support vector machines, decision tree, and deep learning, over score-based models in predicting patient mortality. These models are built to leverage the clinical records gathered during the initial hours after the patient’ admission to the ICU, which are typically inadequate due to time delay in procuring these records from the time-consuming laboratory tests. 

In this study, we will leverage vital signs (specifically heart rate signal) to predict in-hospital mortality. Our research suggests that vital signals can provide numerous information which has been proven to possess a strong relation with mortality [21]. Therefore, vital signal fluctuations can provide high capability to predict the mortality risk more accurately and faster than clinical-based methods.

The goal of this study is to repeat and improve the study done by Harutyunyan et. al. on in-hospital mortality by implementing logistic regression, a standard LSTM, and a channel-wise LSTM on the time series data. This is a binary classification problem, and the success metric will be ROC-AUC. We propose an approach to extract both statistical and signal-based features from the heart signals and employ well-known classifiers (e.g. LSTM and decision tree) to predict in-hospital mortality.
 
Our study focuses on predicting mortality using features extracted from the heart signals of patients within the first forty-eight (48) hours of ICU admission. Quantitative features computed based on the heart rate signals of ICU patients suffering cardiovascular diseases, have been leveraged to predict the risk. Each signal is constructed using 12 statistical and signal-based features. The extracted features are input into eight classifiers: decision tree, linear discriminant, logistic regression, support vector machine (SVM), random forest, boosted trees, Gaussian SVM, and K-nearest neighborhood (K-NN). A few experiments were conducted to observe the performance of these classifiers against the MIMIC-III database.

\section{Related Work}

Harutyunyan et al. (1) proposed a heterogeneous multitask learning problem involving simultaneous learning of four prediction tasks and have created different sets of baselines across various research models. They have outlined a few best practices such as Attention based Recurrent neural networks (LSTM) which are used to identify correlations among prediction tasks. They use a single convolutional network to perform natural language processing tasks (part-of-speech tagging, named entity recognition, and language modeling) with diverse sequential structure. The current project proposed by this team will be to predict in-hospital mortality prediction based on the first 48 hours of an ICU stay. This is a binary classification task with area under the receiver operating characteristic (AUC-ROC) being the main metric. We plan to predict this using the time series of events for the 33,798 unique patients with a total of 42,276 ICU stays.
 
Sadeghi et al. (2) focus on mortality prediction utilizing the MIMIC-III dataset and incorporate features extracted from the heart signals of patients from the first hour of ICU admission. This paper focuses on the subset of patients in the Matched Subset which contain the richer set of vital signals, and further subsets to the study to the nearly 90\% of these patients who stayed in the coronary care unit. Their feature extraction consists of twelve statistical and signal-based features, including the minimum, maximum, range, skewness, kurtosis, standard deviation, variance, mode, averaged power, and energy spectral density. These features are fed into a range of classifiers, and they obtained an AUC score of 0.93 for a decision tree classifier.
 
Xu et al. (3) propose a new model for integrating heterogeneous discrete clinical events and continuous vital signs using an efficient attention mechanism on a sequential deep learning model. The team focused on two tasks: (1) detecting decompensation in an ICU visit as a binary classification problem of whether a patient will die in the next 24 hours, and (2) forecasting the length of stay at the ICU as a multiclass classification problem: days 1-7, 8 days to 14 days, and longer than 14 days. They leverage multi-channel high-density signal processing based on large-scale vital sign recordings including ECG and blood oxygen. They use the MIMIC-III Waveform Database Matched Set,[1] which includes 22,317 waveform records, 22,247 numerics records, and 10,282 clinical discrete records. They define a step size to be one hour, and use an observation length of twelve steps, and utilize batch normalization, ReLU activation and max pooling between convolutional layers, and SAME padding in the model. Their key contribution is multimodal input processing with a guidance matrix from lab measurements and interventions and multi-channel attention mechanism on the processed multimodal input streams. On the decompression prediction, they obtained an AUC-ROC of 90.18\%, and on length of stay, they obtained an accuracy of 86.82\%.
 
Edward Choi et al. (4) propose an RNN-based model that can learn efficient patient representation from a large amount of longitudinal patient records and predict patient diagnosis, medications and length of stay. The dataset used is from Sutter Health Palo Alto Medical Foundation involving 263706 patients. Preprocessing includes grouping ICD9 codes into higher order categories and excluding patients with less than two visits. The multi labeled input of almost 40000 dimensions is projected into a lower dimensional space and then passed as input to the Gated Recurrent Unit implementation of RNN. For predicting diagnosis and medication code, a Softmax hidden layer and ReLU hidden layer for the length of stay are used. The other unique part of the implementation is the use of Skip-gram embedding as an input to the GRU. The performance prediction used is a top-k recall approach. This model achieved 79.58
 
Lipton et al. (7) similarly apply a recurrent neural network to ICU clinical medical data. They presented the first application of LSTMs to multivariate time series of clinical measurements to classify 128 diagnoses from 13 irregularly sampled clinical measurements: systolic and diastolic blood pressure, peripheral capillary refill rate, end-tidal CO2, fraction of inspired O2, Glascow coma scale, blood glucose, heart rate, pH, respiratory rate, blood oxygen saturation, body temperature, and urine output. The authors utilized EHR records from Children’s Hospital LA. They also incorporated additional diagnostic labels by using 301 additional labels from the patient’s chart as a multitask learning approach. They obtained an AUC of 0.8075 and F1 of 0.1530, beating existing MLP implementations at the time.
 
Ghassemi et al, (5) propose combining the latent variable model of extracting free-text clinical notes to features and baseline structured features including the standard physiological measurements from clinical exams, as a single source of data that will be leveraged by models to predict the ICU mortality. The dataset used is MIMIC II 2.6 database with a total of 26,870 patients of Beth Israel Deaconess Medical Center. The input data consists of 50 dimensional vectors of clinical notes for each patient and 36 structured clinical variable feature matrix is extracted. The linear kernel SVM is used for this Classification problem and trained with 30

Desautels et al. (6) apply InSight, a machine learning classification system that uses multivariable combinations of easily obtained patient data like vitals, peripheral capillary oxygen saturation etc.  to predict sepsis. The dataset used is MIMIC-III, of 19828 patients in intensive care unit (ICU) restricted to age 15 years or more. This is a classification problem and the data is partitioned into 4 folds – 3 for training and 1 for testing. The feature selection includes vital sign variables as continuous nonlinear approximations discretized to a temporal frequency of 1 hour and the training uses elastic net regularization, which induces a degree of sparsity among the feature weights. The AUROC of 0.8799 was better than that of Sequential Organ Failure Assessment (SOFA - 0.772) or systemic inflammatory response syndrome (SIRS - 0.609).
 
Zhengping Che et al. (8) propose a Recurrent Neural Network based approach to build deep learning models in cases where there are several missing values in a multivariate time series dataset. The dataset used is MIMIC III which had 58000 hospital admission records from which 99 time series features were extracted. The model was trained on data from the first 48 hours after admission to perform mortality prediction and ICD-9 diagnosis prediction tasks. The approach used was to build a model based on Gated Recurrent Units (GRU-D) with decay rates for each of the input variables and the hidden states. The decay rates are learnt with the training data for each variable and the rate vector is built. An AUC of 0.8527 was achieved compared to 0.7589 for mortality prediction without GRU-D.

\section{MIMIC-III Clinical Data}

The MIMIC-III database with 46,520 patient records has been utilized for this study. This dataset contains variables including vital signs such as blood pressure, heart rate, respiratory rate along with timestamps, lab results, medications, prescriptions, procedures and the beginning and end of interventions. Due to the de-identification process, there are only 10282 patients whose clinical data in the MIMIC-III are associated with the related vital signals in the Matched Subset.
As shown in the Figure 1, the age distributions of the whole MIMIC-III (without infants) and the “Matched Subset” are similar. Hence, the outcomes of the Matched Subset can be extended to the whole database. It is worth mentioning that due to the de-identification process, all the patients greater than or equal to 90 years of age are assigned to one group. Our analysis of the database suggests that nearly 90 percent of patients in the Matched Subset suffer from cardiovascular diseases and are admitted to the coronary care unit (CCU) - an ICU that takes patients with cardiac conditions required continuous monitoring and treatment. Hence, the focus of this study is on predicting the risk of mortality for patients who are admitted to the CCU.

A copy of MIMIC-III data is exported into a local Postgres database. It is a relational database with 26 tables. Tables are linked by identifiers such as, SUBJECT\_ID (unique patient), HADM\_ID (unique admission to the hospital), and ICUSTAY\_ID (unique admission to an intensive care unit). The CHARTEVENTS table contains notes, laboratory tests, and fluid balance. The OUTPUTEVENTS table contains measurements related to patient output, and the LABEVENTS table contains patient laboratory test results. ADMISSIONS, PATIENTS, ICUSTAYS, SERVICES and TRANSFERS tables define and track patient stays (7).

\section{MIMIC-III Waveform Matched Subset}

Our study utilizes the MIMIC-III Waveform Matched Subset, which contains 22,317 waveform records and 22,247 numeric records, which includes periodic measurements of heart rate, oxygen saturation, blood pressure (systolic, mean, and diastolic) and waveforms. These recordings include ECG signals, arterial blood pressure, respiration and photo-plethysmogrom signals. Due to the de-identification process, the matched subset has only 10,282 patient records.

The MIMIC Waveform data was copied to an AWS instance with sufficient storage to accommodate the 2.4 terabytes of Waveform data. All data for a single patient have been stored in a single SUBJECT\_ID subdirectory within one of the ten intermediate-level directories (matched/p00 to matched/p09). Each matched waveform record is of the form matched/pXX/pXXNNNN/pXXNNNN-YYYY-MM-DD-hh-mm, where XXNNNN is the matching MIMIC-III Clinical Database Subject\_ID, and YYYY, MM, DD, hh, and mm are the surrogate year, month (01-12), and day (01-31), and the real hour (00-23) and minute (00-59), derived from the starting date and time of day of the record. The surrogate dates match the corresponding MIMIC-III Clinical Database records. In most cases, the waveform record is paired with a numeric record, which has the same name as the associated waveform record, with an n added to the end. Multiple record pairs are stored chronologically.

\section{Signal Pre-Processing}

Undefined or zero value records will be truncated, and the missing values will be replaced with previously known values. A smoothed heart rate signal S’(t) will be computed using a moving average filter with window size $\rho$ and using the below equation where the original signal S(t) contains L samples.

Since the heart signals were recorded with different lengths and sampling rates of 1 to 0.17 Hz in the MIMIC-III database. Additionally, an anti-aliasing finite impulse response (FIR) low-pass filter [22] will be performed over the low sampling rate signals to avoid bias while comparing signals. The noise samples will be removed by applying the moving average over the original signal. Then, the oversampling method will increase the frequency of the heart rate signal to 1Hz, leading to an increase in the number of samples.

\section{Feature Extraction}

Our study processes the HR signals, and computes a defined set of quantitative features for each signal that are used as the basis for predicting the patient mortality. These quantitative features include twelve (12) statistical and signal-based features which were extracted from the patient’s ECG signal. The statistical features reveal useful information about the distributions of the signal processing data described above. The max, min, and range can demonstrate the spectrum of the signal distribution. The skewness indicates whether the distribution is symmetric or skewed. The kurtosis measures the thickness of the tails of the distribution and the standard deviation shows how the data samples scatter around the mean. The table below indicates the average of each feature for both alive and deceased patients. The reported values indicate the capability of these features in segregating the two groups of patients based on the proposed statistical and signal-based features.

The signal-based features fall into groups like averaged power and power spectral density and are computed as follows:
The averaged power of a finite discrete-time signal is defined as the mean of the signal’s energy. The averaged power of a discrete-time signal S[n] is computed as:

where n1 and n2 are the first and last samples, respectively. The signal power is computed by taking the integral of the power spectral density (PSD) of a signal over the entire frequency space. The PSD is the Fourier transform of the biased estimate of the autocorrelation sequence. The PSD of the signal S[n] with sampling rate $\rho$, in the interval $\Delta$T can be computed as follow

In the MIMIC III dataset, the number of deceased patients inside the hospital is relatively small compared to the alive patients, suggesting that the dataset is imbalanced. The ratio of physiological signals point to the deceased patients in contrast to those who survive is equal to 7:3. Thus, the early mortality prediction systems are faced with an imbalanced dataset. In order to mitigate this issue of imbalance in the dataset, a resampling method of adaptive semi-unsupervised weighted oversampling (A-SUWO) was used to balance the dataset.
 
To reproduce the baseline results, we replicated the methodology described by Harutyunyan et al [1]. The complete MIMIC-III clinical dataset was downloaded locally. The data were reorganized on a per-patient basis, the data were validated, spurious rows were removed, split into episodes, split into train and test sections, and then feature extraction was performed. The pipeline used Python, Numpy, Pandas, and Keras with a Tensorflow backend. Once this was completed, three separate models were trained: logistic regression, a standard LSTM, and a channel-wise LSTM. Performance from these models is shown below.
 
We have implemented the data engineering and feature generation on the MIMIC Waveform data on an AWS cluster using PySpark. 

\section{Experimental Results}

While raw waveform provides tremendously rich detail and potentially significant improvements in model performance, processing and training models on this scale of data poses multiple challenges, including computational complexity, building efficient IO frameworks to accelerate training on large datasets, and vanishing gradients. To make this problem more tractable, we applied multiple feature engineering techniques on the raw waveform data to condense useful information into a compact representation that can be combined with the existing features derived from the clinical datasets in the literature. 

For the logistic regression models, the waveform features were concatenated with the existing clinical features. For the LSTM models, the computation graph was changed to accept both sequential and static features. In our implementation, we pass the waveform features through a single dense hidden layer before merging the outputs with the output of the LSTM on the time-series clinical data, then the combined outputs are jointly passed through a final dense layer.

\begin{table*}[h]
\begin{center}
\begin{tabular}{|l|l|l|l|l|l|l|c|l|}
\hline
Source & Model & AUC-ROC & AUC-PR \\
\hline
Ours & Logistic Regression & 0.849 & 0.486 \\
Harutyunyan et. al & Logistic Regression & 0.848 & 0.474 \\
Percent Difference & Logistic Regression & +0.12\% & +2.53\% \\
\hline
Ours & Standard LSTM & 0.899 & 0.663 \\
Harutyunyan et. al & Standard LSTM & 0.855 & 0.485 \\
Percent Difference & Standard LSTM & +5.15\% & +36.70 \% \\
\hline
Ours & Channelwise LSTM & 0.837 & 0.781\\
Harutyunyan et. al & Channelwise LSTM & 0.862 & 0.515 \\
Percent Difference & Channelwise LSTM & +8.70\% & +51.65\% \\

\hline
\end{tabular}
\caption{AUC-ROC and AUC-PR by model architecture and with waveform features}
\label{tab:results}
\end{center}
\end{table*}

\section{Conclusion}

Hospitals can leverage early mortality prediction to make timely medical decisions and potentially improve clinical outcomes. Our study combines features derived from high-frequency waveform data with traditional clinical for early mortality prediction, leveraging the benefits of statistical and signal-based features from the waveform dataset as well as the diagnoses, prescriptions, and laboratory tests from the clinical dataset. 

In our experiments, we were able to exceed the existing baselines for in-hospital mortality prediction for both logistic regression and deep learning models. For the logistic regression model, our experiments beat the baseline by 0.12\% relative on ROC-AUC and by 2.53\% relative on area under the precision-recall curve. While these gains are incremental, the LSTM especially benefited from the addition of the waveform signals. For the standard LSTM, we observed a 4.09\% increase in ROC-AUC, and a substantial improvement of 31.75\% on area under the precision-recall curve. 

There are multiple potential extensions to this study. First is increasing patient coverage of our waveform features. We decided to focus initially on a single waveform, namely electro-cardiogram, for which records existed for 13.2\% of patients in the MIMIC III dataset. Applying this methodology to additional waveform channels and, consequently, additional patients, would likely improve performances. 

Second is applying this feature extraction technique to multiple time windows. Currently the statistical features are extracted from the ECG time series as a whole, which loses much of the temporal nature of the recordings. A natural extension would be to calculate these features on multiple segmentations of the time series. 

The improvement to prediction quality observed by incorporating waveform features on a subset of the patients suggests significant potential to improve prediction accuracy by increasingly leveraging the rich waveform data that is increasingly available in health care systems. We look forward to seeing additional progress in this area and hope to see it contribute to improving outcomes in the clinical setting.

\end{document}